\documentclass[acmtog, nonacm]{acmart}
\renewcommand\footnotetextcopyrightpermission[1]{} 
\makeatletter
\renewcommand\@formatdoi[1]{\ignorespaces}
\makeatother

\setcopyright{none}
\usepackage{wrapfig}
\usepackage{multirow}
\usepackage{soul}

\begin{document}

\title{Neural Octahedral Field: Octahedral Prior for Simultaneous Smoothing and Sharp Edge Regularization}

\author{Ruichen Zheng}
\email{ankbzpx@hotmail.com}
\orcid{0009-0008-3657-351X}
\affiliation{%
  \institution{Tsinghua University}
  \city{Haidian}
  \state{Beijing}
  \country{China}
}
\affiliation{%
  \institution{Shenzhen University}
  \city{Shenzhen}
  \state{Guangdong}
  \country{China}
}

\author{Tao Yu}
\authornote{Corresponding author: Tao Yu (ytrock@mail.tsinghua.edu.cn).}
\email{ytrock@mail.tsinghua.edu.cn}
\orcid{0000-0002-3818-5069}
\affiliation{%
  \institution{Tsinghua University}
  \city{Haidian}
  \state{Beijing}
  \country{China}
}

\author{Ruizhen Hu}
\email{ruizhen.hu@gmail.com}
\orcid{0000-0002-6798-0336}
\affiliation{%
  \institution{Shenzhen University}
  \city{Shenzhen}
  \state{Guangdong}
  \country{China}
}

\begin{abstract}
Neural implicit representation, the parameterization of a continuous distance function as a Multi-Layer Perceptron (MLP), has emerged as a promising lead in tackling surface reconstruction from unoriented point clouds. In the presence of noise, however, its lack of explicit neighborhood connectivity makes sharp edges identification particularly challenging, hence preventing the separation of smoothing and sharpening operations, as is achievable with its discrete counterparts. In this work, we propose to tackle this challenge with an auxiliary field, the \emph{octahedral field}. We observe that both smoothness and sharp features in the distance field can be equivalently described by the smoothness in octahedral space. Therefore, by aligning and smoothing an octahedral field alongside the implicit geometry, our method behaves analogously to bilateral filtering, resulting in a smooth reconstruction while preserving sharp edges. Despite being operated purely pointwise, our method outperforms various traditional and neural implicit fitting approaches across extensive experiments, and is very competitive with methods that require normals and data priors. Code and data of our work are available at: https://github.com/Ankbzpx/frame-field.
\end{abstract}

\begin{teaserfigure}
    \includegraphics[width=\textwidth]{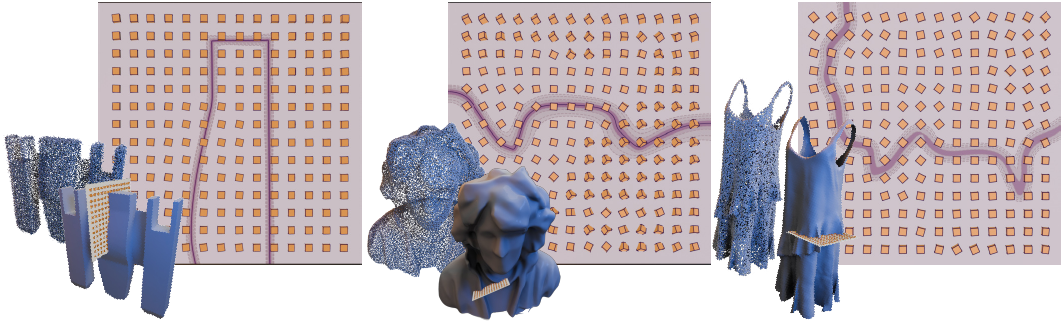}
    \vspace{-15pt}
  \caption{We use the symmetry of octahedral frames to simultaneously smooth and regularize the sharp edges of the 3d reconstructions from unoriented point clouds. The first two are constructed using the signed distance field (SDF), while the last one uses the unsigned distance field (UDF). }
  \Description{This is the teaser figure for the article.}
  \label{fig:teaser}
\end{teaserfigure}

\maketitle
\thispagestyle{empty}

\section{Introduction}

As the predominant method for acquiring real-world geometry, 3D reconstruction technologies have advanced rapidly and become increasingly accessible over the past several decades. The early procedure relied on triangulation in a strictly controlled environment, but now handheld scanning with time-of-flight sensors is no longer a rare sight. In either case, the reconstruction procedure involves acquiring a raw point cloud from multi-view observations, followed by fitting an implicit geometry for final surface extraction. With the development of deep learning and automatic differentiation, parameterizing the implicit surface as a coordinate MLP becomes a popular choice \cite{park2019deepsdf, atzmon2020sal}. Recently, the advancement of neural rendering allows the reconstruction of implicit geometries directly from images for novel view synthesis \cite{mildenhall2021nerf}. Instead of being an intermediate processing format, neural implicit representation has gradually evolved into a direct medium for both producing and consuming 3D geometries, so its direct processing holds practical value.

Since real-world sensors are noise-prone, most surface reconstruction algorithms aim to remove as much noise as possible while preserving geometric features such as sharp edges. In fact, the smoothing and sharpening tasks are often interconnected. For example, \citet{10.1145/1073204.1073227} employ the median filter to remove the initial noise, and then fit a robust subset gradually towards the edges. \citet{10.1145/2421636.2421645} first apply bilateral filtering to denoise off-edge observations and then progressively sample toward the edge to improve its fidelity. Similarly, \citet{https://doi.org/10.1111/cgf.14752} employ line processes to perform smoothing only when similarity measurements fall below a tunable threshold to avoid compromising sharp edges. All works above adhere to the local view of geometry, where spatial structures are captured and communicated through k-Nearest Neighbors (KNN) connectivity on an explicit point cloud.

However, when it comes to neural implicit representation, things are quite different. The typical approach is to encode the geometry as the zero level set of a distance field parameterized by an MLP, so that the geometry can only be inferred and modified through pointwise queries. Although formulating smoothing and sharpening as pointwise energy is not hard \cite{yang2021geometry}, the challenging part is where to apply which---in the pointwise view, there is no local neighborhood connectivity, so traditional edge detection algorithms are no longer applicable. Even if connectivity can be created using differentiable surface extraction \cite{remelli2020meshsdf}, its backpropagation violates the level set equation and therefore is inherently biased \cite{mehta2022level}. Moreover, explicit edge detection is usually multi-staged, which does not fit well with an end-to-end implicit fitting pipeline.

\begin{figure}[]
    \centering
    \includegraphics[width=\linewidth]{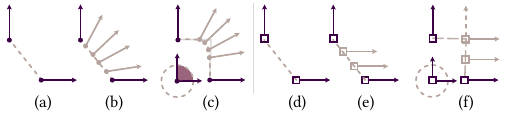}
    \vspace{-20pt}
    \caption{Geometric intuition. (a) Consider a 2D smooth vector field evaluated at two spatial locations as a pair of orthogonal vectors. Due to spatial smoothness, the vectors associated with points in-between need to cover $90^\circ$ angle difference (b). If the vector field is the gradient field of a 2D SDF, and those two points are its zero level samples, then the underlying surface crossing them must have normal vectors covering at least $90^\circ$ on the Gauss map (visualized as solid color on the circle) (c). Without additional constraints or priors, it is difficult to reduce the neighborhood for the normal rotation to produce a visually sharp turning. (d) Now consider a special vector field, the 2D cross field, that exhibits $90^\circ$ symmetry. Since both vectors are equivalent under orthogonality, there is no angle difference between them (e). Therefore, when used as a guidance for the SDF gradient field, the cross field naturally induces a sharp geometric prior (f).}
    \label{fig:intuition}
    \Description{Intuition}
\end{figure}

In this work, we propose a method that gives a uniform treatment of both smoothing and sharpening the neural implicit representation, which also stays pointwise throughout the whole process. Our key observation is based on how the octahedral frames--the 3D generalization of cross frames \cite{10.1145/3084873.3084921, 10.1111/cgf.12014} and originally designed for hex meshing \cite{BEOctahedral}--are interpolated across edges. Their symmetry allows for a sharp jump in directional constraints (see Fig. \ref{fig:intuition}), making them naturally edge-aware. Hence, both smoothing and sharp edge regularization can be reduced to smoothing in octahedral space.

To this end, we pair the distance field with spatially varying octahedral frames parameterized by an additional MLP, which we refer to as the \emph{neural octahedral field}. By aligning with the distance gradient and enforcing smoothness in octahedral space, it serves as a prior for simultaneously smoothing and sharpening the underlying geometry. Compared to explicit edge detection and alternating between smoothing and sharpening losses, our method offers a uniform treatment for the two objectives. More importantly, our method remains pointwise throughout the process, making it a great fit for neural implicit reconstruction tasks.

Our contributions can be summarized as:
\vspace{-4pt}
\begin{itemize}
    \item Model spatially varying octahedral frames as a neural implicit field and derive a novel pointwise loss to align it with the distance field;
    \item Utilize the symmetry of octahedral frames to regularize the gradient of a distance field, hence smoothing while simultaneously emphasizing sharp turnings across the edges;
    \item Integrate our method for both signed and unsigned distance field surface reconstruction techniques.
\end{itemize}

\begin{figure*}[t]
    \centering
    \includegraphics[width=\linewidth]{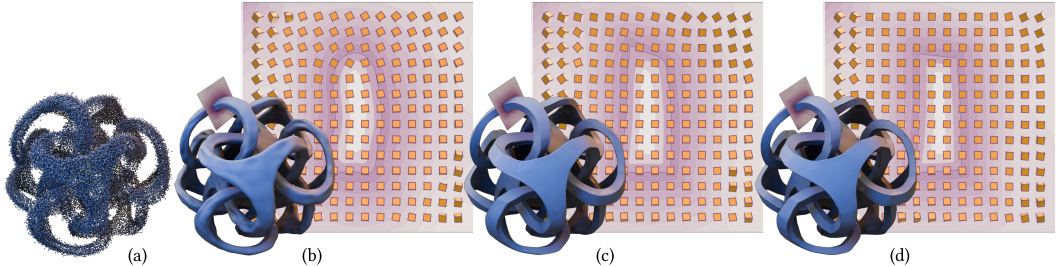}
    \vspace{-15pt}
    \caption{The illustration of our pipeline. Given an unoriented point cloud (a), we initialize a distance field. By pairing with an aligned octahedral field and encouraging smoothness in octahedral space (b-d), we perform simultaneous smoothing and sharpening of the underlying geometry.}
    \label{fig:pipeline}
    \Description{optimization process.}
\end{figure*}

\section{Related work}

\paragraph{Surface reconstruction from noisy point cloud}
Surface reconstruction has been extensively researched in the last decades and continues to evolve rapidly with the advancement of deep learning. We focus on reviewing approaches most related to ours and refer the interested readers to \cite{huang2022surface, sulzer2024survey} for more detailed reviews.

The point set surface (PSS) \cite{964489} fits polynomials to approximate the surface locally. Algebraic Point Set Surface (APSS) \cite{10.1145/1276377.1276406} fits the algebraic sphere to improve the stability of PSS over regions of high curvature. The variable implicit point set surface (VPSS) \cite{10.1145/3306346.3322994} leverages the eikonal constraint to ease the need for input normals, but at the cost of cubic complexity. In contrast to local fitting, Poisson Surface Reconstruction (PSR) \cite{kazhdan2006poisson} represents the surface globally as the level set of an implicit indicator function, whose gradient matches the input normals. Screened Poisson Surface Reconstruction (SPSR) \cite{kazhdan2013screened} integrates positional constraints to balance precision and smoothness. Recently, Neural Kernel Surface Reconstruction (NKSR) \cite{huang2023nksr} integrates aspects from both ends--it models the surface as the zero level set of a global implicit function, but with its value determined by the learned proximity (neural kernel) with respect to input samples.

\paragraph{Patch-based Point cloud resampling and denoising}
The majority of surface reconstruction methods treat the point cloud normals as the noisy gradient of a local polynomial or implicit function, such that they are well adapted to handling noisy directional constraints. However, their presumptions often fall short for positional noise that needs to be treated as outliers and excluded via robust filtering \cite{https://doi.org/10.1111/j.1467-8659.2009.01388.x} or forward search \cite{10.1145/1073204.1073227}. Therefore, it is a common practice to resample and denoise point clouds in advance.

Edge-Aware Point Set Resampling (EAR) \cite{10.1145/2421636.2421645} applies bilateral filtering to reliable fit normals away from edges, then progressively samples toward the edge to capture sharp features. RFEPS \cite{Xu_2022} utilizes the local planar assumption of CAD models to perform aggressive denoising and feature lines extraction. \citet{8490996} randomly sample square patches, store those local height values in square matrices, then perform denoising via low-rank matrix factorization. \citet{zeng20193d} sample overlapping patches and connect samples by projection distance to perform Graph Laplacian Regularization (GLR). \citet{https://doi.org/10.1111/cgf.14752} use line processes (LP) to perform smoothing up to an optimizable threshold. \citet{Williams_2019} parameterize local patches with MLPs and rely on their smoothness prior for denoising. \citet{wei2023ipunetiterative} predict cross-field at local patches and leverage its crease alignment nature \cite{jakob2015instant, huang2016extrinsically} to form a spatial tessellation. They then rasterize the point cloud for iterative upsampling via 3D convolution. IterativePFN \cite{de_Silva_Edirimuni_2023} embeds an iterative module into the network structure to learn noise filtering and patch stitching.

\paragraph{Fitting based Neural implicit representation}
Neural implicit representation encodes the surface as the zero-level set of a continuous distance function using a coordinate MLP. We primarily review its application in fitting-based surface reconstruction. Readers interested in the neural field in general should check out the comprehensive survey by \citet{DBLP:journals/corr/abs-2111-11426}.

DeepSDF \cite{park2019deepsdf} and Occupancy Networks \cite{OccupancyNetworks} parameterize the signed distance function (SDF) or occupancy classification over spatial locations, which are capable of encoding geometry of arbitrary resolutions. SAL \cite{atzmon2020sal} samples in the vicinity of the input point cloud and learns an unsigned distance function. SALD \cite{atzmon2021sald} further improves the reconstruction quality with normal supervision. IGR \cite{icml2020_2086} introduces the eikonal regularization term, which encourages MLPs to converge to faithful SDFs under stochastic optimization. When normals are not available, the eikonal regularization is unstable \cite{yang2023steikstabilizingoptimizationneural}, hence yielding ambiguities and artifacts. \citet{park2023ppoisson} model the SDF gradient as the unique solution to the $p$-Poisson equation and additionally minimize the surface area to improve hole filling. DiGS \cite{benshabat2023digs} initializes the MLP with a geometric sphere and minimizes the divergence of its gradient to preserve the orientation consistency. Neural Singular Hessian (NSH) \cite{wang2023neuralsingularhessian} encourages the Hessian of MLP to have zero determinant and produces a topologically faithful initialization. StEik \cite{yang2023steikstabilizingoptimizationneural} uses the equivalent constraint as NSH, but formulated as Hessian gradient products, which greatly improves the stability of fitting. NeurCADRecon (NeurCAD) \cite{Dong2024NeurCADRecon} minimizes Gaussian curvature to reconstruct developable surfaces. However, since zero Gaussian curvature is not topologically feasible everywhere, they introduce a double-trough function to encourage $\pi / 2$ curvature as secondary minimal and gradually anneal loss weight during the fitting process.

\section{Method}
Given an unoriented point cloud $\{x_i\}_{i=1}^N$, we fit an initial distance field $u(x)$ parameterized by an MLP, with its zero level set encoding the reconstructed surface:
\begin{equation}
S = \{ x \in \mathbb{R}^3 | u(x)=0 \}.
\end{equation}
Similarly, we assign every spatial point in $\mathbb{R}^3$ an octahedral frame, which we denote as the octahedral field $V(x)$. By being smooth while aligning with the distance gradient $\nabla u(x)$ as much as possible, the octahedral field serves as a prior for simultaneously smoothing and regularizing sharp edges of the level set surface. Our full pipeline is illustrated in Fig. \ref{fig:pipeline}.

\subsection{Definition of Neural Octahedral Field}
\label{sec:octa-parameterization}
We can think of an octahedral frame $V$ as the assignment of three mutually orthogonal directions $v_1, v_2, v_3$ and their opposites at a given point:
\begin{equation}
    V = \{\pm v_1, \pm v_2, \pm v_3\}.
\end{equation}

To model a smooth octahedral field, we enforce that for two spatially adjacent points $x_i, x_j \in \mathbb{R}^3$, the difference between their associated octahedral frames $V(x_i)$ and $V(x_j)$ is small. To this end, we need a parameterization to measure the difference between two octahedral frames. A natural choice is their representation vectors, with three sufficient to characterize a frame.
However, due to cubic symmetry, permuting or negating those vectors does not modify the frame they represent. For example, $v_2, -v_1, v_3$ and $-v_1, v_2, v_3$ indicate the same frame. In fact, there are 24 combinations, and measuring the difference between two frames requires the representation vectors matching \cite{SHOctahedral}. Therefore, it is more convenient to design a parameterization that is invariant under such symmetry. 

\paragraph{Symmetry-invariant Parameterization}
\citet{SHOctahedral} parameterize an octahedral frame $V$ functionally as the following polynomial integrated over the unit sphere $\mathcal{S}^2 = \{s \in \mathbb{R}^3 \ | \ \|s\| = 1\}$:
\begin{equation}
    F'_V(s) = (v_0 \cdot s)^2 (v_1 \cdot s)^2 + (v_1 \cdot s)^2 (v_2 \cdot s)^2 + (v_2 \cdot s)^2 (v_0 \cdot s)^2.
\end{equation}
It is evident that the functional value of $F'_V(s)$ is invariant to the permutation and negation of $v_i$. For subsequent derivation, we use the equivalent form proposed by \citet{ray2016practical}, denoted as $F_V$
\begin{equation}
    \label{eq:functional-polynomial-explicit}
    F_V(s) = 1 - 2F'_V(s) = \sum_{i=1}^3 (v_i\cdot s)^4, \ s \in \mathcal{S}^2.
\end{equation}
\citet{SHOctahedral} further losslessly project the polynomial onto the Spherical Harmonics (SH) band 0 and 4 basis as follows:
\begin{equation}
    \label{eq:functional-polynomial-sh}
    F_q(s) = c_0 (c_1 y_0 + q^T y_4(s)), s \in \mathcal{S}^2, \|q\| = 1,
\end{equation}
where $c_0, c_1 \in \mathbb{R}$ are constants, $y_0 \in \mathbb{R}$ is SH band 0 basis, $y_4(s) \in \mathbb{R}^9$ is the vector form of SH band 4 basis evaluated at $s$, and $q \in \mathbb{R}^9$ is the associated band 4 coefficient vector.

For the octahedral frame $V_a$ and $V_b$, let $F_{q_a}$ and $F_{q_b}$ be functional polynomials parameterized by $q_a$ and $q_b$ respectively. Due to the orthogonality of the SH basis, the difference between two frames can be reduced to the difference between their SH band 4 coefficient vectors \cite{ray2016practical}:
\begin{equation}
    \label{eq:smoothness-derivation}
    \begin{aligned}
        d(V_a, V_b) =&  \int_{\mathcal{S}^2} (F_{q_a}(s) - F_{q_b}(s))^2 ds \\ &\propto \int_{\mathcal{S}^2} (q_a^T y_4(s) - q_b^T y_4(s))^2 ds \\
        &= (q_a - q_b)^T (\int_{\mathcal{S}^2} y_4(s) y_4(s)^T ds) (q_a - q_b) \\
        &= \|q_a - q_b\|_2^2.
        \end{aligned}
\end{equation}

Therefore, an octahedral frame can be fully characterized by its associated SH band 4 coefficient vector $q$. Following \citet{algebraic}, we refer to the set of $q$, representing all octahedral frames, as the octahedral variety.

\paragraph{MLP Representation}
We can then associate every spatial location $x \in \mathbb{R}^3$ with an octahedral frame parameterized by the SH coefficient vector $q$. That is, we can think of $q$ as a function of $x$ and represent it using an MLP, which we refer to as our \emph{neural octahedral field}:
\begin{equation}
    q: \mathbb{R}^3 \to \mathbb{R}^9.
\end{equation}

Note that since the band 4 coefficient vectors have unit norm, we normalize the MLP output to limit its solution space. It is worth mentioning that the unit norm is only a necessary condition--without further constraints, the output of our MLP is not guaranteed to represent valid octahedral frames.

\subsection{Alignment with Distance Field}
Given a neural octahedral field $q$, we aim to align it with the gradient of the distance field $u$, that is, one of the representation vectors $v_i$ of $q(x)$ needs to match the direction of $\nabla u(x)$ for $x \in \mathbb{R}^3$. 

To update $q$ via backpropagation, we need to measure the misalignment of $q(x)$ with normalized $\nabla u(x)$, or more generally, with any direction vector $r \in \mathbb{R}^3$. The immediate option is to use their dot product $v_i \cdot r$, but to do so, we need a way to recover the representation vector $v_i$ from $q$. 
We will first introduce several algebraic characteristics of octahedral frames and then derive the corresponding alignment conditions.

\paragraph{Algebraic characteristics}
\citet{algebraic} characterize the set of octahedral frames as a subset of 4-th order symmetric orthogonally decomposable tensor $T$ of the following form:
\begin{equation}
    T = \sum_{i=1}^3 v_i^{\otimes 4},
\end{equation}
where $\otimes$ denotes tensor power, and representation vectors $v_i$ of octahedral frame are the eigenvectors of $T$ with eigenvalues 1.
Moreover, the one-to-one corresponding homogeneous polynomial of $T$ has the following form:
\begin{equation}
    \label{eq:homogeneous-polynomial}
    F_T(r) = \sum_{i=1}^3 (v_i\cdot r)^4, r \in \mathbb{R}^3.
\end{equation}

When comparing Equ. (\ref{eq:functional-polynomial-explicit}) and (\ref{eq:homogeneous-polynomial}), we can find that the functional polynomial $F_V$ of an octahedral frame is exactly its associated homogeneous polynomial $F_T$ restricted to $\mathcal{S}^2$, that can be recovered as:
\begin{equation}
    F_T(r) = |r|^4 F_V(\frac{r}{|r|}), r \in \mathbb{R}^3.
\end{equation}
Since the SH projection is lossless, $F_T$ can be equivalently parameterized with $q$ as
\begin{equation}
    F_T(q, r) = |r|^4 F_q(\frac{r}{|r|}) = c_0 (c_1 \hat{y}_0(r) + q^T \hat{y}_4(r)),
\end{equation}
where $\hat{y}_0(r) = |r|^4 y_0$, $\hat{y}_4(r) = |r|^4 y_4(\frac{r}{|r|})$.

The set $V$ is exactly the set of robust eigenvectors of its orthogonally decomposable tensor $T$, which are also the fixed points of its homogeneous polynomial \cite{Robeva_2016}. As such, they can be recovered using the tensor power method \cite{Lathauwer1995HigherorderPM}:
\begin{equation}
    r_{t} = \frac{\nabla_r F_T(q, r_{t-1})}{\|\nabla_r F_T(q, r_{t-1})\|}, \lim_{t \to \infty} r_t = v_i.
\end{equation}
However, as an iterative algorithm, it is difficult to propagate gradients back to $q$, so we instead need a more automatic differentiation-friendly formulation.

\paragraph{Optimization-friendly formulation} \label{sec:align-condition}
\citet{Robeva_2016} identifies that a general vector $r \in \mathbb{R}^3$ is the eigenvector of $T$ if and only if the gradient of its homogeneous polynomial evaluated at $r$ is collinear with $r$:
\begin{equation}
    \label{eq:homogeneous-polynomial-gradient}
    \nabla F_T (r) = 4 r.
\end{equation}
Therefore, Equ. (\ref{eq:homogeneous-polynomial-gradient}) is also a necessary and sufficient condition for $r$ to be a representation vector of an octahedral frame $V$. We can then rewrite the equation using $q$ as:
\begin{equation}
    \label{eq:alignment-condition}
    \nabla_r F_T(q, r) = 4 r.
\end{equation}
Note that Equ. (\ref{eq:alignment-condition}) is a highly compact and bidirectional constraint. If $q$ represents a valid octahedral frame, $r$ satisfies Equ. (\ref{eq:alignment-condition}) must be one of its representation vectors, so $r$ implicitly satisfies $|r| = 1$. Conversely, if $r$ is a unit vector, $q$ satisfies Equ. (\ref{eq:alignment-condition}) must be an $r$-aligned octahedral frame, so it implicitly forces $q$ to lie on the octahedral variety.

\begin{figure*}[]
    \centering
    \includegraphics[width=\linewidth]{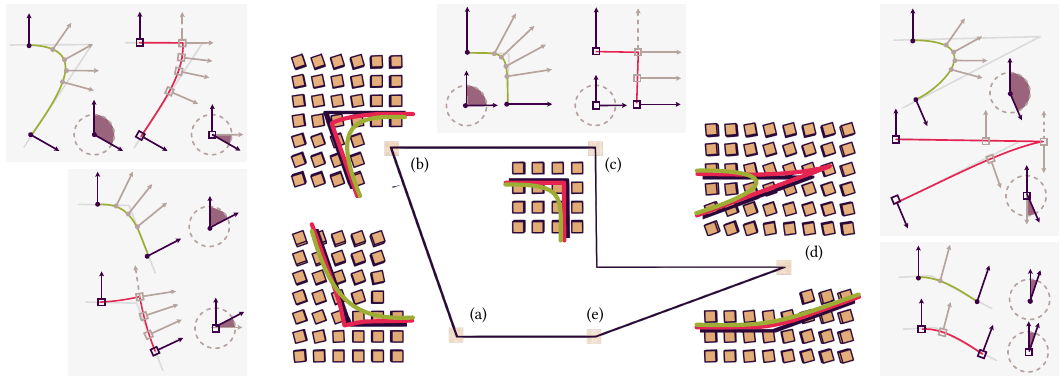}
    \vspace{-20pt}
    \caption{Non-$90^\circ$ dihedral angle handling, the green strokes indicate the zero level sets under smooth vector interpolation, while the red strokes indicate the ones under our guidance. Although $90^\circ$ symmetry cannot precisely describe a non-$90^\circ$ angle, it can still serve as a prior for visually sharp features (a-d), and when the angle difference is small, it converges to smooth vector interpolation (e).}
    \label{fig:cases}
    \Description{Cases}
\end{figure*}

\subsection{Optimization}
As shown in Fig. \ref{fig:pipeline}, our method adopts a two-stage pipeline--after initialization of the distance field, we fit a smooth and aligned octahedral field, which simultaneously guides the update of the distance field gradient till convergence.
To this end, we define two new losses, $\mathcal{L}_{\text{smooth}}(q)$ that encourages smoothness in the octahedral field $q$, and $\mathcal{L}_{\text{align}} (q, u)$ that aligns $q$ with the distance field $u$. Our total losses are
\begin{equation}
    \begin{aligned}
\mathcal{L}_{\text{total}}(q, u) &= \tau (\lambda_{\text{smooth}} \mathcal{L}_{\text{smooth}}(q) + \lambda_{\text{align}} \mathcal{L}_{\text{align}} (q, u)) 
\\
&+ \mathcal{L}_{\text{recon}} (u),
\end{aligned}
\end{equation}
where $\tau$ is a linear weight scheduling factor to shift stage, $\lambda_{\text{smooth}}$ and $\lambda_{\text{align}}$ are smooth and align loss weights, $\mathcal{L}_{\text{recon}}$ refers to the collection of losses of corresponding initialization method. 

It is worth mentioning that, although the octahedral frame can only describe angles that are multiples of $90^\circ$, our guidance remains effective for general angles, since the octahedral field still provides a sharper prior than smooth vector interpolation (Fig. \ref{fig:cases}). Also note that, unlike traditional multi-stage methods that shift focus from off-edge to close-edge, or even shift representation from local patch to implicit surface, we retain pointwise perception, stay implicit, and optimize in an end-to-end fashion.

\paragraph{Smoothness loss}
To encourage smoothness of the octahedral field, we minimize the continuous form of Equ. (\ref{eq:smoothness-derivation}), which is the classic Dirichlet energy:
\begin{equation}
    \label{eq:smooth-loss}
    \mathcal{L}_{\text{smooth}} (q) =
\int_{\mathbb{R}^3} w_u(x) \|\nabla q (x)\|_F^2 dx.
\end{equation}

\paragraph{Alignment loss} \label{sec:align-loss}
To align the gradients of the distance field with octahedral frames, we rewrite Equ. (\ref{eq:alignment-condition}) as a distance measurement:
\begin{equation}
    \label{eq:normalize}
    d(q, u) = \|\nabla_r F_T(q, r) - 4 r\|_2^2, r = \frac{\nabla u}{|\nabla u|}.
\end{equation}
Note that $F_T$ is analytical, so $\nabla_r F_T$ can be computed in closed form. The normalization of $\nabla u$ yields two benefits: First, it decouples the scaling of $|\nabla u|^3$ when evaluating the gradient of $\nabla_r F_T$ over $q$, hence stabilizing the training; Additionally, it disentangles the unit norm constraint so our method is applicable for non-unit norm distance fields. 
The alignment loss is then defined as:
\begin{equation}
    \mathcal{L}_{\text{align}} (q, u) = \int_{\mathbb{R}^3} w_u(x) d(q(x), u(x)) dx,
\end{equation}
where $w_u(x) = \exp (-\beta \cdot  SG(u(x)))$ is a weight function that prioritizes samples close to the zero level set and $SG$ is the stop gradient operator.

\paragraph{Reconstruction loss}
Our method can be paired with any reconstruction method as long as it initializes a faithful distance field from the unoriented point cloud. Note that $\mathcal{L}_{\text{recon}}$ is still needed after initialization, as our method is by no means conservative--particularly the normalization of $\nabla u$ (Sec. \ref{sec:align-loss}) can cause its norm to shrink indefinitely if no further constraints are imposed.

\begin{figure*}[t]
    \centering
    \includegraphics[width=\linewidth]{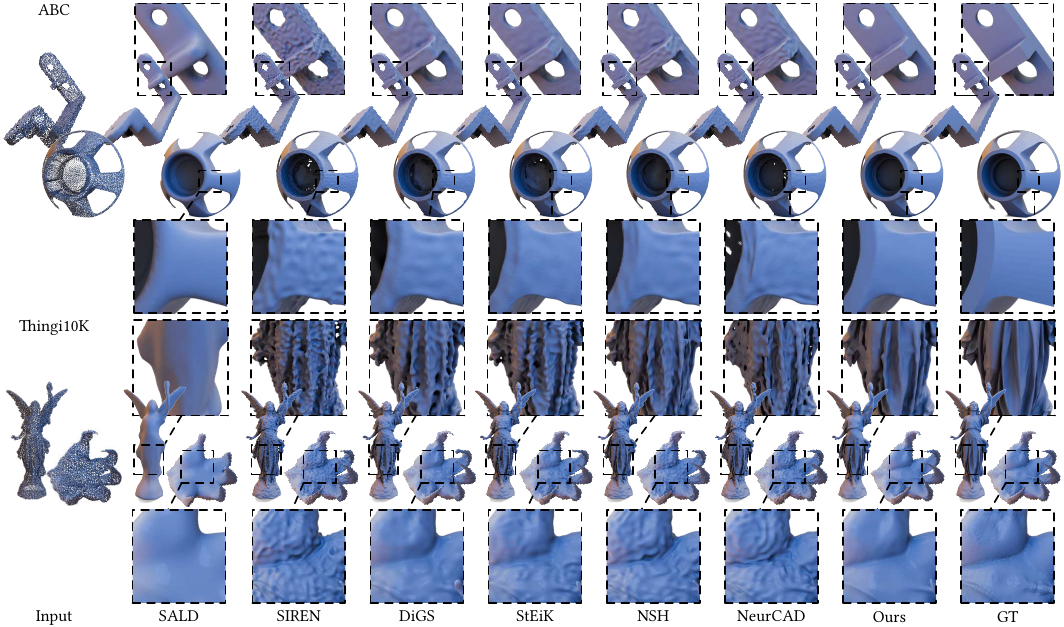}
    \vspace{-15pt}
    \caption{Qualitative results of implicit fitting methods on ABC / Thingi10k at noise level $0.002L$. Our method outperforms others with cleaner shape features.}
    \label{fig:implicit}
    \Description{implicit}
\end{figure*}

\section{Experiments and Results}
In this section, we present a comprehensive comparison with a broader range of object-level surface reconstruction methods that use noisy point clouds as input. Since SDF is the most common neural implicit representation for this task, our evaluation primarily focuses on SDF-based reconstruction, as detailed in Section~\ref{sec:sdf}. 
To further demonstrate the flexibility of our method, we integrate it with unsigned distance fields (UDF) and perform additional comparisons in Section~\ref{sec:udf}.


\subsection{Experiment Setup}
\paragraph{Implementation Details}
We model both distance field and octahedral field using a 4-layer SIREN \cite{sitzmann2019siren} of 256 units, with input normalized to $[-1, 1]$. Since the octahedral field needs to keep up with the update of the distance field, we set $\lambda_{\text{align}}$ to match the weight scaling of the initialization method. We set $\beta=100$ \cite{ma2023better} and fix relative weight $\lambda_{\text{smooth}} = 0.01 \cdot \lambda_{\text{align}}$. For each scan, we optimize using the ADAM optimizer \cite{kingma2017adammethodstochasticoptimization} with a learning rate of $5 \times 10^{-5}$ over 10k iterations on a single RTX 3090 GPU of 24GB VRAM, which takes around 12 minutes.

We implement our method using JAX \cite{jax2018github} (Equinox \cite{kidger2021equinox} for neural networks, optax \cite{deepmind2020jax} for optimizers). Our implementation is available at: https://github.com/Ankbzpx/frame-field.

\begin{table}[]
    \caption{Comparisons to neural implicit fitting methods under noise level $0.002L$. The bold text indicates the best scores, and the underlined text indicates the second best. }
    \vspace{-4pt}
    \label{tab:implicit}
    \begin{tabular}{lccc}
        \toprule
         & \multicolumn{3}{c}{ABC / Thingi10k ($0.002L$)}           \\
         & \multicolumn{1}{c}{Chamfer $\downarrow$}        & \multicolumn{1}{c}{Hausdorff $\downarrow$}             & \multicolumn{1}{c}{F-score $\uparrow$} \\ \hline                                                                                                                                                                                                                                                        \\ 
        SALD
        &12.434 / 6.737 & 13.059 / 8.446 & 64.044 / 67.129                                                                                                                                                                                                                                                                \\
        SIREN
         & 8.835 / 11.019 & 14.805 / 18.854 & 87.537 / 85.593                                                                                                                                                                                                                                                              \\
        DiGS
         & \underline{3.734} / \underline{2.232} & 10.640 / 9.124 & \underline{93.885} / \underline{96.775}                                                                                                                                                                                                                                                           \\
        StEik
         & 4.128 / 2.611 & \underline{7.099} / 6.530 & 92.084 / 95.207                                                                                                                                                                                                                                                               \\
        NSH
         & 5.755 / 3.792 & 8.847 / \underline{6.356} & 92.391 / 96.064                                                                                                                                                                                                                                                                \\
        NeurCAD
         & 5.447 / 4.531 & 9.090 / 9.817 & 91.732 / 94.529                                                                                                                                                                                                                                                                    \\
        Ours
         & \textbf{2.663} / \textbf{2.038} & \textbf{5.495} / \textbf{3.880} & \textbf{94.325} / \textbf{97.078}                                                                                                                                              \\
        \bottomrule
    \end{tabular}
    \vspace{-10pt}
\end{table}

\paragraph{Evaluation Metrics}
For quantitative evaluation, we follow DiGS \cite{benshabat2023digs} and report the Chamfer distance $(\times 10^3)$ \cite{fan2016point} and the Hausdorff distance $(\times 10^2)$ over 1M points randomly sampled on the surface or from the point cloud, depending on the output of the methods. Given that neural implicit representations are susceptible to floating artifacts, we additionally report the F-score \cite{Knapitsch2017}. It clamps the closest matching distance by a threshold and reports it as a percentage, hence is less sensitive to the mismatching of surplus parts. Following \citet{what3d_cvpr19}, we use the distance threshold of $0.5\%$. Note that we do not postprocess the output for any methods and leave the extracted meshes or point clouds as is.

For point cloud visualization, we use SPSR \cite{kazhdan2013screened} if the method outputs normals (EAR \cite{10.1145/2421636.2421645}), otherwise we use the Advancing Front \cite{ZIENKIEWICZ2013573} (GLR \cite{zeng20193d}, LP \cite{https://doi.org/10.1111/cgf.14752}, IterativePFN \cite{de_Silva_Edirimuni_2023}).

\subsection{SDF Reconstruction}
\label{sec:sdf}


\begin{figure*}[t]
    \centering
    \includegraphics[width=\linewidth]{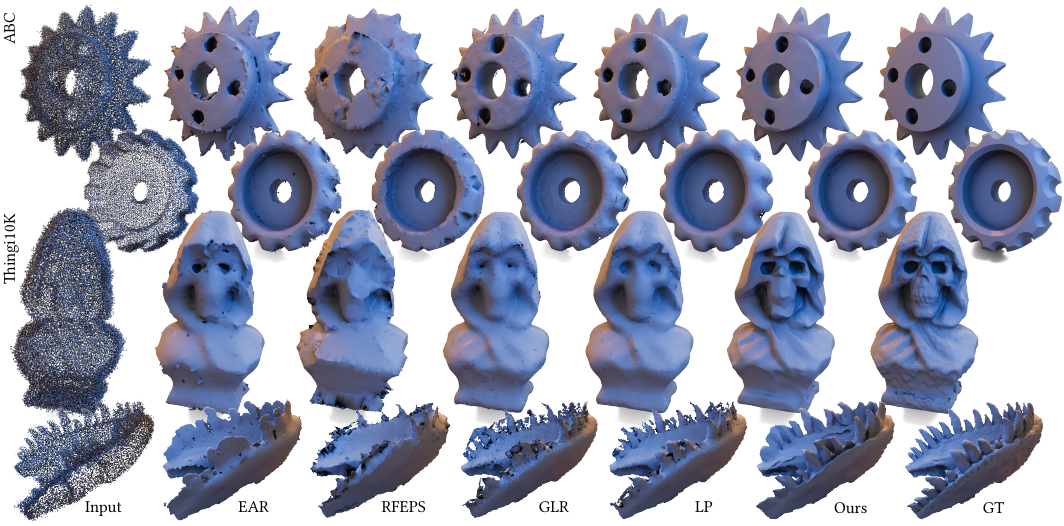}
    \caption{Qualitative result of denoising methods on ABC / Thingi10k. Our method performs better at highlighting sharp edges and maintaining organic shape features.}
    \label{fig:denoise}
    \Description{Denosing}
\end{figure*}

\paragraph{Baselines.}
We compare our method with 14 different methods over four categories:
\vspace{0.05cm}\\
$\bullet$ \textit{6 implicit fitting ones}: SALD \cite{atzmon2021sald}, SIREN \cite{sitzmann2019siren}, DiGS \cite{benshabat2023digs}, StEik \cite{yang2023steikstabilizingoptimizationneural}, NSH \cite{wang2023neuralsingularhessian}, NeurCAD \cite{Dong2024NeurCADRecon}. Similar to all methods in this category, we also fit a continuous SDF parameterized by a single MLP.
\vspace{0.05cm}\\
$\bullet$ \textit{4 patch denoising ones}: EAR \cite{10.1145/2421636.2421645}, RFEPS \cite{Xu_2022}, GLR \cite{zeng20193d}, LP \cite{https://doi.org/10.1111/cgf.14752}. Methods in this category represent traditional multi-stage pipelines, in which points within patches are smoothed, while points between distinctly oriented patches are resampled or projected to preserve sharp edges. Our method, on the other hand, offers an end-to-end, uniform treatment for both smoothing and sharpening.
\vspace{0.05cm}\\
$\bullet$ \textit{2 axiomatic ones}: APSS \cite{10.1145/1276377.1276406}, SPSR \cite{kazhdan2013screened}. APSS approximates the implicit surface at local regions, while SPSR fits a global discretized implicit function. In contrast, our methods fit a global continuous implicit function, but update locally through pointwise energy.
\vspace{0.05cm}\\
$\bullet$ \textit{2 data prior ones}: NKSR \cite{huang2023nksr}, IterativePFN \cite{de_Silva_Edirimuni_2023}. Although our method employs a neural network, it relies on geometric priors of octahedral frames rather than those learned from data.


\paragraph{Datasets.}
We perform extensive comparisons on two widely used datasets: ABC \cite{Koch_2019_CVPR} and Thingi10k \cite{zhou2016thingi10k}. The former consists of CAD models of sharp edges, whereas the latter is made up of more general shapes. 
We follow NKSR \cite{huang2023nksr}, use the 100 test splits from Points2Surf \cite{Erler_2020} and Blensor \cite{gschwandtner2011blensor} to simulate the time-of-flight (ToF) scanning process. Specifically, we set sensor resolution to $176 \times 144$, focal length to 10 mm, and scanning each object spherically for 30 scans, resulting in point clouds of size ranging from 20k to 100k. We generate scanning data for two noise levels, $\mathcal{N}(0, 0.002L)$ and $\mathcal{N}(0, 0.01L)$ (so $400$ in total), with both noise added to sensor depth, where $L$ is the length of the maximum edge of the model's bounding box. For reference, NKSR \cite{huang2023nksr} randomizes two noise levels $0.01L$ and $0.05L$--we set the noise level to their lower bound, as a higher level is difficult to handle without data prior.

\begin{table}[h]
    \caption{Comparisons to patch denoising methods under noise level $0.01L$. The bold text indicates the best scores, and the underlined text indicates the second best. 
    }
    \vspace{-4pt}
    \label{tab:denoise}
    \begin{tabular}{lccc}
        \toprule
         & \multicolumn{3}{c}{ABC / Thingi10k ($0.01L$)}           \\
         & \multicolumn{1}{c}{Chamfer $\downarrow$}        & \multicolumn{1}{c}{Hausdorff $\downarrow$}             & \multicolumn{1}{c}{F-score $\uparrow$} \\ \hline                                                                                                                                                                                                                                                        \\ 
        EAR
         &\underline{4.375} / 3.843 & \underline{6.133} / 4.393 & \underline{81.057} / 78.195                                                                                                                                                                                                                                                                 \\
        RFEPS
         &16.744 / 8.270 & 13.174 / 7.833 & 54.829 / 54.115                                                                                                                                                                                                                                                                   \\
        GLR
         &4.965 / \underline{3.603} & 6.233 / \textbf{3.756} & 75.153 / \underline{79.139}                                                                                                                                                                                                                                                                \\
        LP
         & 5.917 / 4.514 & \textbf{6.132} / \underline{3.810} & 60.472 / 61.982 \\
         Ours
         & \textbf{4.145} / \textbf{3.237} & 7.534 / 5.258 & \textbf{88.866} / \textbf{91.500} \\
        \bottomrule
    \end{tabular}
    \vspace{-10pt}
\end{table}

\begin{figure*}[t]
    \centering
    \includegraphics[width=\linewidth]{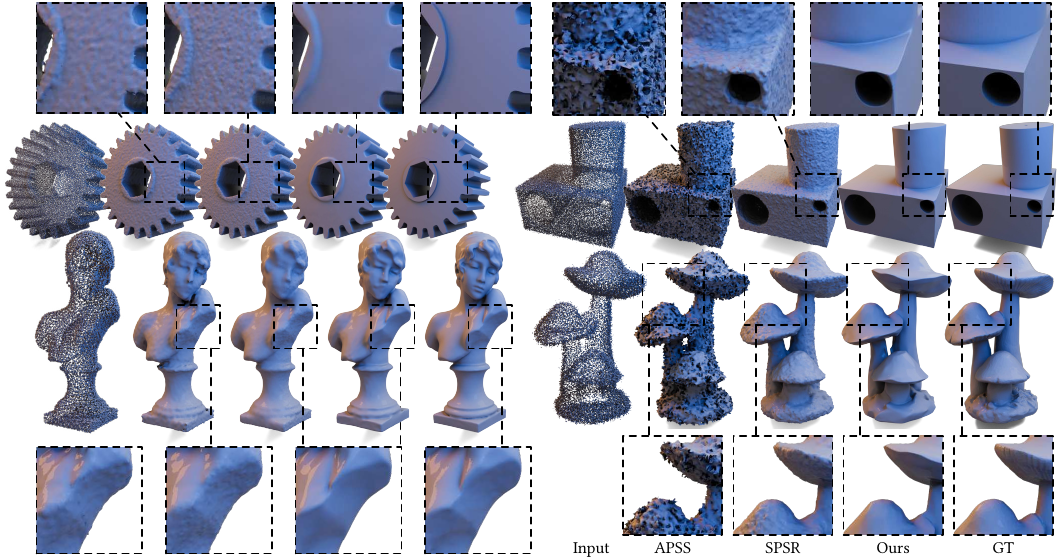}
    \vspace{-15pt}
    \caption{Qualitative results of axiomatic methods. The top row is from ABC, and the bottom row is from Thingi10k. The left column is with noise level $0.002L$, and the right column is with noise level $0.01L$. Our method produces cleaner flag regions and sharper tunings.}
    \label{fig:axiomatic}
    \Description{Axiomatic ones}
\end{figure*}

\begin{figure*}[h]
    \centering
    \includegraphics[width=\linewidth]{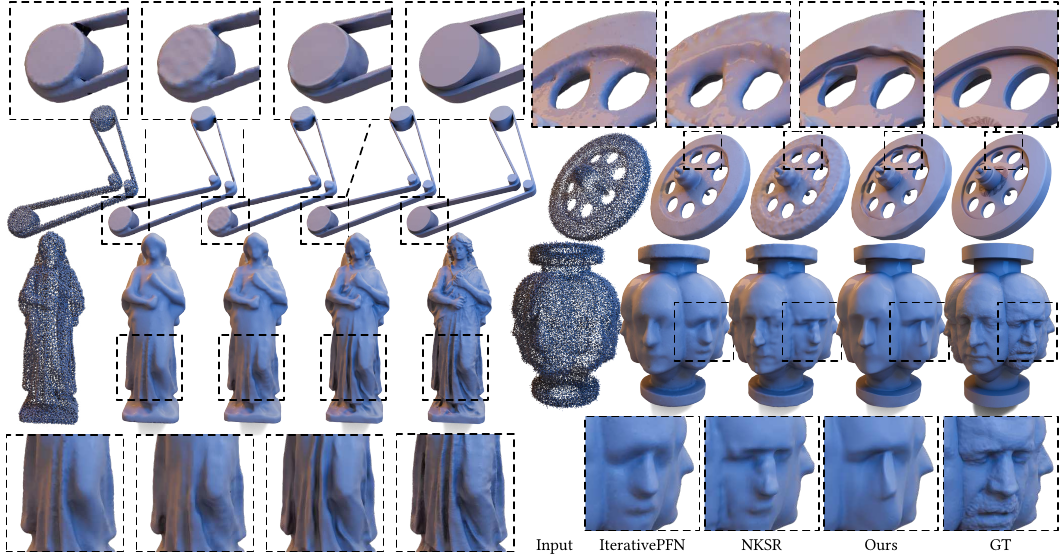}
    \caption{Qualitative results of methods with data priors. The top row is from ABC, and the bottom row is from Thingi10k. The left column is with noise level $0.002L$, and the right column is with noise level $0.01L$. Our method inherently lacks data prior that cannot handle structures at different scales and may introduce distortion when the noise level is high. }
    \label{fig:data-prior}
    \Description{Data prior}
\end{figure*}

\paragraph{Reconstruction details.} We use NSH \cite{wang2023neuralsingularhessian} as our initialization method. For noise level $0.002L$, we use NSH official loss weights. For noise level $0.01L$, since their default weights are not tuned to handle large noise, we tune them accordingly. We leave details of initialization tuning in the supplementary material. For both noise levels, we set $\lambda_{\text{align}} =50$ to match their weight of Eikonal term and $\lambda_{\text{smooth}} = 0.5$. We further extract the surface using the Marching Cube (MC) with a voxel grid of resolution $512^3$, the common practice of various implicit fitting work \cite{benshabat2023digs, yang2023steikstabilizingoptimizationneural, wang2023neuralsingularhessian}.

\paragraph{Comparison results.}
For clarity, we compare with the baselines category by category. The full metrics are provided in the supplementary material. 

For comparison with \emph{implicit fitting methods}, including SALD \cite{atzmon2021sald}, SIREN \cite{sitzmann2019siren}, DiGS \cite{benshabat2023digs}, StEik \cite{yang2023steikstabilizingoptimizationneural}, NSH \cite{wang2023neuralsingularhessian}, NeurCAD \cite{Dong2024NeurCADRecon}, we use their officially released code and keep their tuned weights as is. We extract all surfaces using MC of resolution $512^3$ and highlight our comparison at noise level $0.002L$ (Fig. \ref{fig:implicit}, Tab. \ref{tab:implicit}), where we do not adjust initialization weights. Therefore, cross-referencing the reconstructions gives a fair assessment of our intrinsic smoothing and sharpening capability. Our method outperforms others, by not only producing cleaner feature lines for CAD models, but also handling general features such as palmar creases and cloth foldings better. Unsurprisingly, DiGS performs the second best, as its smoothness prior suppresses the noise, but at the cost of losing sharp features.

For comparison with \emph{patch denoising methods}, we use the EAR \cite{10.1145/2421636.2421645} implemented in CGAL \cite{10.1145/1653771.1653865} with PCA normal (KNN $k=16$), filtered by scanning view directions. For RFEPS \cite{Xu_2022}, GLR \cite{zeng20193d}, LP \cite{https://doi.org/10.1111/cgf.14752}, we use their official implementations. All methods are tuned according to their instructions, with their parameters documented in the supplementary material. For fairness, we highlight our comparison at noise level $0.01L$ (Fig. \ref{fig:denoise}, Tab. \ref{tab:denoise}), as all methods are designed to handle large noise. Compared to EAR and RFEPS, our method performs better in maintaining organic shapes. Compared to GLR and LP, our method produces sharper edges.

For comparison with \emph{axiomatic methods}, including APSS \cite{10.1145/1276377.1276406} and SPSR \cite{kazhdan2013screened}, we use MeshLab \cite{LocalChapterEvents:ItalChap:ItalianChapConf2008:129-136} implementations with PCA normal (KNN $k=16$) and default parameters. Our method consistently produces visually superior results, as illustrated in Fig. \ref{fig:axiomatic}.

For comparison with \emph{data prior methods}, we use the official pretrained models for both NKSR \cite{huang2023nksr} (kitchen-and-sink) and IterativePFN \cite{de_Silva_Edirimuni_2023}. Our method inherently lacks data prior that cannot adapt to structures of different scales. Therefore, if our assumptions are met, we can produce seemingly cleaner reconstructions. However, violating those assumptions may lead to missing details or low-poly distortions (Fig.~\ref{fig:data-prior}).

\subsection{UDF Reconstruction}
\label{sec:udf}
Given that the alignment of octahedral frames is agnostic to the flip and norm scaling of the distance gradient, our method can be applied to more general distance fields, such as unsigned, non-unit norm ones.

\begin{figure*}[h]
    \centering
    \includegraphics[width=0.98\linewidth]{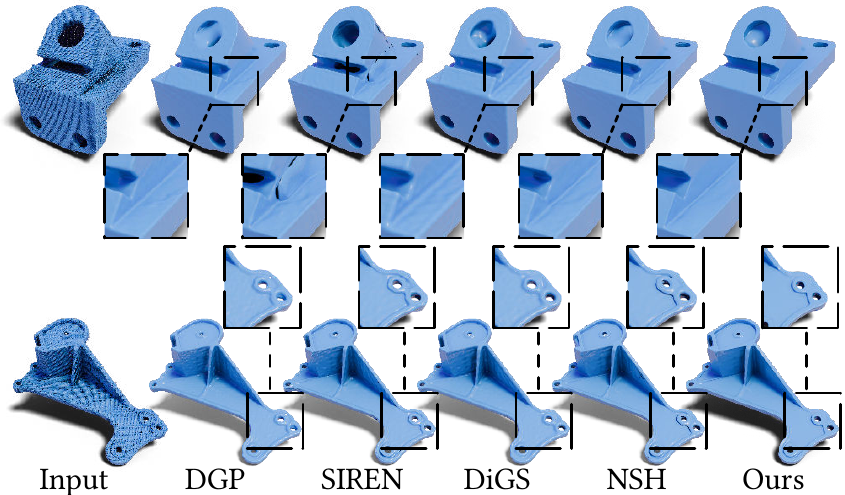}
    \vspace{-10pt}
    \caption{Qualitative results of UDF reconstructions on SRB \cite{berger2013benchmark}. Our method outperforms others with cleaner reconstructions.}
    \label{fig:srb}
    \Description{SRB qualitative results.}
    \vspace{-8pt}
\end{figure*}

\citet{yang2024monge} recently propose to model the (scaled) squared SDF as a standalone distance field, known as the $\text{S}^2\text{DF}$. Unlike typical unsigned distance fields that take the absolute value of SDF, $\text{S}^2\text{DF}$ is differentiable at the zero level set, though with vanishing gradient norm. We apply our method to $\text{S}^2\text{DF}$ \cite{yang2024monge} using Pytorch \cite{paszke2019pytorchimperativestylehighperformance}. We set $\lambda_{\text{align}} = 8 \times 10^{6}$ to match their weight of Neumann term and $\lambda_{\text{smooth}} = 8 \times 10^{4}$. Similarly, we leave the initialization details in the supplementary material.

We additionally compare our method with $\text{S}^2\text{DF}$ \cite{yang2024monge} and CapUDF \cite{zhou2024cappami} on the Surface Reconstruction Benchmark (SRB) dataset \cite{berger2013benchmark}, which consists of 5 scans that exhibit triangulation-based scanning patterns. For fair comparison, we use DCUDF \cite{Hou2023DCUDF} to extract the surface of all three methods with grid resolution $256^3$. Our method achieves the best metrics (Tab. \ref{tab:srb-udf}), which is in alignment with our cleaner reconstruction quality (Fig. \ref{fig:srb}).

\begin{figure}[]
    \centering
    \includegraphics[width=\linewidth]{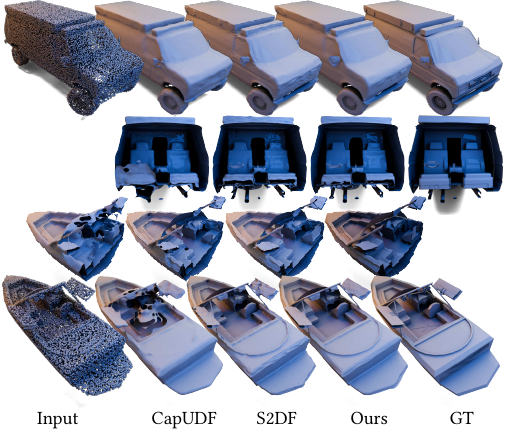}
    \vspace{-20pt}
    \caption{Qualitative results on ShapeNet Car and Watercraft using UDF.}
    \label{fig:udf}
    \Description{UDF results.}
    \vspace{-8pt}
\end{figure}

Given that UDF is known for its capability to represent open surfaces and complex interiors, we additionally pick 20 models each from the Car and Watercraft categories of the ShapeNet dataset \cite{chang2015shapenetinformationrich3dmodel}. To capture the interior, we directly sample 50k points uniformly on each surface to run the comparison. Although performing better quantitatively (Tab. \ref{tab:srb-udf}), our method can slightly extend the open surface (Fig. \ref{fig:udf}). This is limited by our non-conservative nature and the lack of explicit open boundary constraints in the existing reconstruction loss.

\begin{table}[]
    \caption{Quantitative results on SRB \cite{berger2013benchmark} and 40 models of ShapeNet \cite{chang2015shapenetinformationrich3dmodel}. The bold text indicates the best score.}
    \vspace{-4pt}
    \label{tab:srb-udf}
    \begin{tabular}{lcccccc}
        \toprule
              & \multicolumn{3}{c}{SRB / ShapeNet}           \\
              & \multicolumn{1}{c}{Chamfer $\downarrow$} & \multicolumn{1}{c}{Hausdorff $\downarrow$} & \multicolumn{1}{c}{F-score $\uparrow$} \\ \hline
        CapUDF   & 0.218 / 1.550                                                                         & 4.866 / 1.739                                   & 92.116 / 89.817         \\
        $\text{S}^2\text{DF}$ & 0.201 / 0.835                                                  & 4.578 / \textbf{1.395}                                      & 92.853 / 99.013           \\
        Ours (UDF)  & \textbf{0.200} / \textbf{0.808}                                                            & \textbf{4.538} / 2.130                         & \textbf{93.114} / \textbf{99.259}   \\
        \bottomrule
    \end{tabular}
    \vspace{-10pt}
\end{table}

\subsection{Ablation Studies}
We focus on ablating major designs of our method and leave the discussions of alternative loss formulations to the supplementary material.

\begin{figure}[h]
    \centering
    \includegraphics[width=\linewidth]{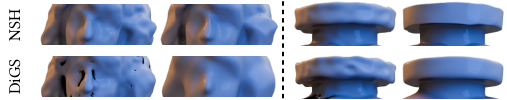}
    \vspace{-15pt}
    \caption{Two constrained cases with different initializations. 
    For each case, the left figure shows the initialization and the right one shows the reconstruction. A smoother initialization makes it harder for the octahedral field to identify sharp edges (left), and smoothing the distance field results in slower emergence of sharp features (right).}
    \label{fig:digs-ablation}
    \Description{Smoothness ablation}
    \vspace{-10pt}
\end{figure}

\paragraph{Initialization method}
Our method pairs better with an initialization that does not introduce additional smoothness prior to the distance field. To demonstrate, we experiment with DiGS as our initialization and observe a performance drop at a higher noise level (Tab. \ref{tab:Initialization}). This is because when the noise level is high, minimizing the Laplacian produces an overly smooth initialization, such that further fitting an octahedral field and smoothing in octahedral space makes it difficult for details to emerge. Moreover, a smooth loss will also dilute the alignment update, making sharp features harder to form (Fig. \ref{fig:digs-ablation}).

\begin{table}[h]
    \caption{Initialization ablation. Explicitly smoothing the distance field makes it difficult to recover sharp edges, resulting in reduced performance.}
    \label{tab:Initialization}
    \begin{tabular}{lccc}
        \toprule
         & \multicolumn{3}{c}{ABC / Thingi10k ($0.01L$)}           \\
         & \multicolumn{1}{c}{Chamfer $\downarrow$}        & \multicolumn{1}{c}{Hausdorff $\downarrow$}             & \multicolumn{1}{c}{F-score $\uparrow$} \\ \hline                                        Ours (DiGS)
         & 4.953 / 4.342 & 8.018 / 7.160 & 81.326 / 77.782 \\
         Ours (NSH)
         & \textbf{4.145} / \textbf{3.237} & \textbf{7.534} / \textbf{5.258} & \textbf{88.866} / \textbf{91.500} \\
        \bottomrule
    \end{tabular}
    \vspace{-10pt}
\end{table}

\begin{figure}[h]
    \centering
    \includegraphics[width=\linewidth]{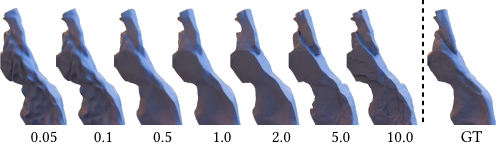}
    \vspace{-15pt}
    \caption{We fix the $\lambda_{\text{align}}=50$ while increasing the $\lambda_{\text{smooth}}$ from $0.05$ to $10.0$ (from left to right). As smoothness increases, the initial noisy structure becomes smoother and sharper, but a larger $\lambda_{\text{smooth}}$ can distort reconstruction, resulting in missing details and unwanted sharp edges.}
    \label{fig:smoothness-ablation}
    \Description{Smoothness ablation}
    \vspace{-10pt}
\end{figure}

\paragraph{Smoothness weight}
The smoothing and sharpening effect of the octahedral field is governed by the relative weight between $\mathcal{L}_{\text{align}}$ and $\mathcal{L}_{\text{smooth}}$. Fixing $\lambda_{\text{align}}$ while increasing $\lambda_{\text{smooth}}$ will cause sharp features to emerge. However, an overly large $\lambda_{\text{smooth}}$ would remove a large portion of the alignment constraints, resulting in misleading updates on the distance field gradient, hence distorting the reconstruction (Fig. \ref{fig:smoothness-ablation}). We empirically find $\lambda_{\text{smooth}}=0.5$, that is, $\lambda_{\text{smooth}}=0.01 \cdot \lambda_{\text{align}}$ to be a robust trade-off across the whole dataset, which we keep fixed throughout our experiments.

\begin{figure}[]
    \centering
    \includegraphics[width=\linewidth]{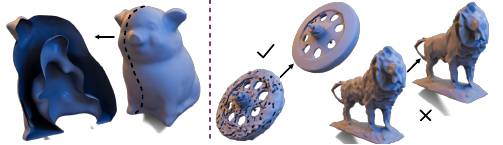}
    \vspace{-10pt}
    \caption{Our method is prone to ghost geometries and surplus parts (left, using SDF to represent an open hole), and cannot fix a corrupted initialization (right, DiGS and NSH initialization respectively).}
    \label{fig:limitation1}
    \Description{Update ablation}
\end{figure}

\begin{figure}[t]
    \centering
    \includegraphics[width=\linewidth]{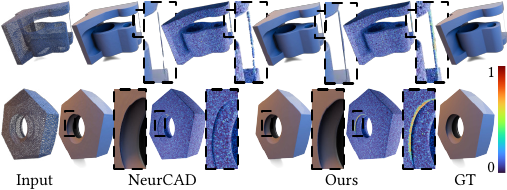}
    \vspace{-20pt}
    \caption{Sharp edge recovery comparison with NeurCAD on noise-free input. The color-coded error map visualizes one-sided Chamfer distance from reconstruction to ground truth, with error normalized by $0.5\%$ of the bounding box diagonal. Our current formation does not scale down for very thin structures, and our prior cannot precisely describe non-$90^\circ$ tips.}
    \label{fig:cad}
    \Description{cad qualitative results.}
    \vspace{-8pt}
\end{figure}

\begin{table}[]
    \caption{Runtime performance of implicit fitting methods (in minutes) and corresponding average Chamfer distance on ABC and Thingi10K at noise level 0.002L.}
    \label{tab:runtime}
    \small
    \begin{tabular}{lccccccc}
        \toprule
         & \multicolumn{1}{c}{SALD}        & \multicolumn{1}{c}{SIREN}             & \multicolumn{1}{c}{DiGS} & \multicolumn{1}{c}{StEik} & \multicolumn{1}{c}{NSH} & \multicolumn{1}{c}{NeurCAD} & \multicolumn{1}{c}{Ours} \\ \hline                                        Runtime
         & 2.5 & 4 & 7.5 & 13.5 & 7 & 7 & 12
 \\
         Chamfer
         & 9.586 & 9.927 & 2.983 & 3.370 & 4.774 & 4.989 & 2.351
 \\
        \bottomrule
    \end{tabular}
    \vspace{-10pt}
\end{table}

\section{Conclusion}
We introduce the neural octahedral field, which, when paired with neural implicit representation, can serve as priors for simultaneously smoothing while emphasizing sharp edges of embedded geometry. We design a mathematically clean and effective loss that aligns those two fields with pointwise evaluation. We extensively compare our method with existing baselines to demonstrate our effectiveness in both SDF and UDF reconstruction tasks. In future work, we would like to explore the integration of the neural octahedral field with data prior for adaptive scaling.

\paragraph{Limitations and future work}
Our method requires a faithful initialization, as smoothing in octahedral space cannot resolve a fully corrupted geometry. Given that the octahedral frame is invariant to the sign flip of the directional constraints, it also cannot fix the inconsistent normal orientation as with NSH. Similar to other neural implicit fitting approaches, our method is prone to ghost geometries and surplus parts (Fig. \ref{fig:limitation1}). It is also slower (Tab. \ref{tab:runtime}), as we model the octahedral field using an additional MLP and we minimize its Jacobian norm, a high-order loss (Equ. \ref{eq:smooth-loss}). Moreover, our method is a non-conservative gradient-based filtering approach, which requires other reconstruction losses to bound the geometry. Therefore, when positional constraints are weak, particularly for the boundary of open surfaces (Fig. \ref{fig:udf}), we can slightly extend the reconstruction. Given that the octahedral field cannot precisely describe non-$90^\circ$ angles (Fig. \ref{fig:cad}), future work may consider replacing it with a non-orthogonal frame field. Although our method is motivated by the unique point-wise constraints inherent to MLP-based distance field fitting, integrating our method with other surface representations, particularly those with local neighborhoods or even global information available, is an interesting direction. Finally, our current formulation treats structures at different scales equivalently--an adaptive scaling is needed to apply our method to scene-level reconstruction, such as in neural rendering, with examples provided in the supplementary material. 
\section*{Acknowledgement}
We thank the anonymous reviewers for their valuable comments. This work was supported by the National Key R$\&$D Program of China (No.2024YFB2809101) and NSFC (No.62322207, No.62171255).

\bibliographystyle{ACM-Reference-Format}
\bibliography{neural-octahedral-field}

\clearpage
\appendix
\section{Experiment Details}

\begin{table*}[]
    \caption{The results from two noise levels are separated by a slash, with left indicating noise $\sigma = 0.002L$ and right $\sigma = 0.01L$. Note methods marked with * require normal input (PCA normal with KNN $k=16$, filtered by scanning view directions), and methods marked with '\dag' leverage data prior. The bold text indicates the best scores, while the underlined text indicates the best scores for methods that do not need normal input.}
    \label{tab:full-metrics}
    \begin{tabular}{lccc|ccc}
        \toprule
         & \multicolumn{3}{c|}{ABC \cite{Koch_2019_CVPR}}  & \multicolumn{3}{c}{Thingi10k \cite{zhou2016thingi10k}}                                                                                                                                                                            \\
         & \multicolumn{1}{c}{Chamfer $\downarrow$}        & \multicolumn{1}{c}{Hausdorff $\downarrow$}             & \multicolumn{1}{c|}{F-score $\uparrow$} & \multicolumn{1}{c}{Chamfer $\downarrow$} & \multicolumn{1}{c}{Hausdorff $\downarrow$} & \multicolumn{1}{c}{F-score $\uparrow$} \\ \hline
        APSS* \cite{10.1145/1276377.1276406}
         & \textbf{2.333} / 4.863
         & \textbf{5.043}  / 9.078
         & \textbf{94.693}    / 70.838
         & \textbf{1.346}   / 4.333
         & 3.367   / 7.960
         & \textbf{97.542}   /   68.216                                                                                                                                                                                                                                                        \\
        SPSR* \cite{kazhdan2013screened}
         & 3.306  / 3.999
         & 6.091   / 6.162
         & 91.756  / 89.185
         & 1.892 / 2.765
         & 3.939 / 4.338
         & 96.437 / 93.299                                                                                                                                                                                                                                                                     \\
        EAR* \cite{10.1145/2421636.2421645}
         & 4.066  / 4.375
         & 5.785   / 6.133
         & 84.405  / 81.057
         & 3.590  / 3.843
         & 3.541 / 4.393
         & 80.505 / 78.195                                                                                                                                                                                                                                                                     \\
        RFEPS* \cite{Xu_2022}
         & 17.297  / 16.744
         & 13.375   / 13.174
         & 62.593  / 54.829
         & 6.939  / 8.270
         & 7.707 / 7.833
         & 67.055 / 54.115                                                                                                                                                                                                                                                                     \\
        NKSR* \dag \cite{huang2023nksr}
         & 2.929  / \textbf{3.600}
         & 6.579  /  7.152
         & 93.636 / \textbf{90.184}
         & 1.594  / \textbf{2.338}
         & 5.127 / 6.696
         & 97.082 / \textbf{93.307}                                                                                                                                                                                                                                                            \\ 
         \hline
        GLR \cite{zeng20193d}
         & 4.026  / 4.965
         & 5.768  /  6.233
         & 82.602 / 75.153
         & 2.774 / 3.603
         & 3.236  / \underline{\textbf{3.756}}
         & 87.909 /  79.139                                                                                                                                                                                                                                                                    \\
        LP \cite{https://doi.org/10.1111/cgf.14752}
         & 4.601 / 5.917
         & 5.634  / \underline{\textbf{6.132}}
         & 76.228 / 60.472
         & 3.464 / 4.514
         & \underline{\textbf{3.173}} / 3.810
         & 78.731 / 61.982                                                                                                                                                                                                                                                                     \\
        IterativePFN \dag \cite{de_Silva_Edirimuni_2023}
         & 3.726 / 4.386
         & 6.047  / 6.366
         & 85.960 / 79.658
         & 2.499 / \underline{3.077}
         & 3.426 / 3.757
         & 89.042 / 82.938                                                                                                                                                                                                                                                                     \\ 
        SALD \cite{atzmon2021sald}
         & 12.434  / 16.465
         & 13.059 /   18.017
         & 64.044  / 42.587
         & 6.737 / 15.061
         & 8.446 / 15.064
         & 67.129 / 42.732                                                                                                                                                                                                                                                                     \\
        SIREN \cite{sitzmann2019siren}
         & 8.835  / 6.427
         & 14.805 /   8.978
         & 87.537  / 58.945
         & 11.019 / 5.710
         & 18.854 / 9.071
         & 85.593 / 55.810                                                                                                                                                                                                                                                                     \\
        DiGS \cite{benshabat2023digs}
         & 3.734  / 6.590
         & 10.640 / 11.484
         & 93.885 / 58.463
         & 2.232 / 6.046
         & 9.124 / 8.042
         & 96.775 /  51.056                                                                                                                                                                                                                                                                    \\
        StEik \cite{yang2023steikstabilizingoptimizationneural}
         & 4.128  / 5.819
         & 7.099 / 7.758
         & 92.084 / 64.552
         & 2.611 / 5.682
         & 6.530 / 5.614
         & 95.207 /  57.138                                                                                                                                                                                                                                                                    \\
        NSH \cite{wang2023neuralsingularhessian}
         & 5.755 / 5.420
         & 8.847  / 7.516
         & 92.391 /  64.614
         & 3.792 / 5.119
         & 6.356  / 7.386
         & 96.064  / 59.365                                                                                                                                                                                                                                                                    \\
        NeurCAD \cite{Dong2024NeurCADRecon}
         & 5.447 / 5.169
         & 9.090  / 7.109
         & 91.732 /  66.533
         & 4.531 / 4.463
         & 9.817  / 6.618
         & 94.529  / 62.935                                                                              \\
         Ours (DiGS)
         & \underline{2.637} / 4.953
        & 5.607	/ 8.018
        & \underline{94.400} / 81.326
        & \underline{1.981}	/ 4.342
        & 3.989	/ 7.160
        & 96.938 / 77.782
         
         \\
        Ours (NSH)
         & 2.663  / \underline{4.145}
         & \underline{5.495}  / 7.534
         & 94.325  / \underline{88.866}
         & 2.038  / 3.237
         & 3.880 / 5.258
         & \underline{97.078} / \underline{91.500}                                                                                                                                                   \\
        \bottomrule
    \end{tabular}
\end{table*}

\subsection{SDF Reconstruction}
\paragraph{Initialization}
NSH \cite{wang2023neuralsingularhessian} has 4 losses, $\mathcal{L}_{\text{input}}$ which encourages zero distance value for input samples, $\mathcal{L}_{\text{eikonal}}$ which enforces the eikonal constraint for input samples, $\mathcal{L}_{\text{off}}$ which encourages the large distance value for off-surface samples, $\mathcal{L}_{NSH}$ which regularizes singular Hessian for close-surface samples. Following their official implementation, at each iteration, we draw 15k samples from the input point cloud, 15k off-surface samples from the uniform distribution $\mathcal{U}(-1, 1)$, and 15k close-surface points from normal distributions centered on the input samples, with sigma the maximum KNN distance of $k=51$. We additionally apply $\mathcal{L}_{\text{eikonal}}$ to off-surface samples because we empirically find that it fixes most failed reconstructions of noisy close-to-planar objects.

\begin{figure}[]
    \centering
    \includegraphics[width=\linewidth]{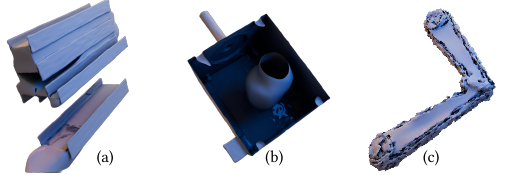}
    \vspace{-15pt}
    \caption{We observe 3 types of ghost geometries that NSH can not resolve: (a) very thin planar structures, (b) distance beyond close samples where NSH loss is applied, (c) after NSH loss anneals and noise accumulates.}
    \label{fig:nsh-Limitations}
    \Description{NSH limitations.}
\end{figure}

The official weights of NSH losses are $\lambda_{\text{input}}=7000$, $\lambda_{\text{off}}=600$, $\lambda_{\text{eikonal}}=50$, $\lambda_{\text{NSH}}$ is initialized to $3$ and then annealed to $0.0001$ after $20\%$ iterations for a period of $20\%$ iterations. For noise level $0.002L$, we find that official weights perform well, and we schedule $\tau$ at $60\%$ for a period of $20\%$ iterations to ensure sufficient details are captured. For noisy level $0.01L$, we lower $\lambda_{\text{input}}$ to $3500$ and annealed $\lambda_{\text{NSH}}$ to $0.001$ to avoid over noise corruption (Fig. \ref{fig:nsh-Limitations}). We then schedule $\tau$ at $40\%$ for the period of $20\%$ iterations. In both noise levels, we set $\lambda_{\text{align}} = 50, \lambda_{\text{smooth}} = 0.5$.

\begin{figure*}[]
    \centering
    \includegraphics[width=\linewidth]{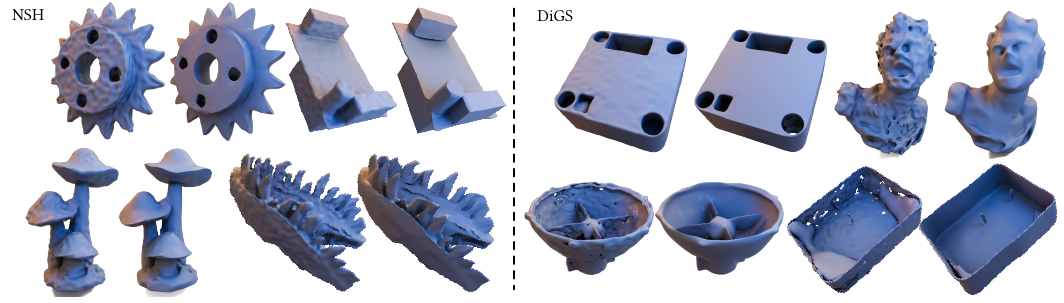}
    \vspace{-15pt}
    \caption{Visualizations of our initialization and final reconstruction for noise level $0.01L$.}
    \label{fig:two-initializations}
    \Description{NSH limitations.}
\end{figure*}

The tuning may appear counterintuitive, as for different noise levels, we adjust $\mathcal{L}_{\text{recon}}$ and $\tau$ scheduling rather than our loss weights. This is because our method essentially functions like bilateral filtering--it cannot add new details or fix fully corrupted parts of the initial geometry. Similar to traditional denoising methods, the relative weight between $\lambda_{\text{align}}$ and $\lambda_{\text{smooth}}$ is the presumption made to determine which parts to smooth and which parts to preserve, which we fix for objects of similar scales. Therefore, we instead tune initialization to suppress large noise levels so our method can perform further filtering (Fig. \ref{fig:two-initializations}).

In ablation of DiGS \cite{benshabat2023digs} initialization, we follow the official code to set initial divergence loss weight to $100$ then anneal it to $0$. We keep the annealing and other loss weights consistent with NSH initialization (except we remove the Hessian term).

\paragraph{Methods in comparison}
We use official implementations of SALD \cite{atzmon2021sald}, SIREN \cite{sitzmann2019siren}, DiGS \cite{benshabat2023digs}, StEik \cite{yang2023steikstabilizingoptimizationneural}, NSH \cite{wang2023neuralsingularhessian}, NeurCAD \cite{Dong2024NeurCADRecon}, as well as their default weights and scheduling for comparison. Note that for NeurCAD \cite{Dong2024NeurCADRecon}, we use code with commit 794a1, 4 hidden layers each with 256 neurons, learning rate $5 \times 10^{-5}$, and 15k on-, close-, off-surface samples per iteration, running for 10k iterations per reconstruction. We use weights of $7000$ for on-surface and $600$ for off-surface Dirichlet constraints, $50$ for Eikonal constraint, $10$ for Gauss loss that linearly drops to $0.001$ from $20\%$ to $50\%$ iterations, and eventually annealed to zero in the end. Their code does not include the double-trough curve, so we implement it faithfully according to their paper.

\paragraph{Patch Denoising}
We use the EAR \cite{10.1145/2421636.2421645} implemented in CGAL \cite{10.1145/1653771.1653865} with PCA normal (KNN $k=16$), filtered by scanning view directions. For RFEPS \cite{Xu_2022}, GLR \cite{zeng20193d}, LP \cite{https://doi.org/10.1111/cgf.14752}, we use their official implementations. All methods are tuned according to their instructions as follows:
\begin{itemize}
    \item[EAR] For the bilateral smoothing stage, we set the sharpness angle to $25^\circ$, iteration number to $3$, KNN $k$ to $50$ or $120$ for two noise levels. For edge-aware upsampling, we set sharpness angle to $25^\circ$, edge sensitivity to $0.3$, neighbor radius to $0.25$, and output sample number multiplier to $4$.
    \item[RFEPS] We set KNN $k$ to $60$ or $120$ for two noise levels. For $0.01L$, we additionally enable the denoising stage.
    \item[GLR] We set the maximum iterations to $20$, termination tolerance to $0.00005$, patch size to $30$, search window size to 16, $\lambda_a=25$, $\lambda_b$ to $4$ or $7$ for two noise levels.
    \item[LP] We set $w_a=1$, $w_b=5000$, $\mu_{\text{smooth}}=30$, lp threshold to $0.3$, $\mu_{\text{fit}}=5\times 10^{-9}$, maximum smoothing iteration to 5. We also set $w_c$ to $0.5$ or $1$, KNN $k$ to $15$ or $50$, for two noise levels accordingly.
\end{itemize}

\subsection{UDF Reconstruction}

\paragraph{Initialization.}
$\text{S}^2\text{DF}$ \cite{yang2024monge} models squared distance field scaled by a constant factor $K$, that we set $K=1000$ according to their suggestions. $\text{S}^2\text{DF}$ has 4 losses, $\lambda_{\text{Dirichlet}}$ which encourages zero distance value for input samples, $\lambda_{\text{Neumann}}$ which enforces vanishing gradient norm for input samples, $\lambda_{\text{Non-manifold}}$ which encourages the large distance value for close-surface samples, $\lambda_{\text{Monge-Ampere}}$ which regularizes its second-order property for both input and close-surface samples. We refer to their official implementation to normalize the input point cloud in the range $[-1, 1]$. For each iteration, we draw 15k samples from the input point cloud and 15k close-surface points from normal distributions centered on the input samples, with sigma $0.1$.

We use their official weights $\lambda_{\text{Dirichlet}}=10^8$, $\lambda_{\text{Neumann}}=8 \times 10^6$, $\lambda_{\text{Non-manifold}}=10^6$, $\lambda_{\text{Monge-Ampere}}=8.5 \times 10^{-3}$. To match their scaling, we set $\lambda_{\text{align}}=8 \times 10^6$ and $\lambda_{\text{smooth}}=8 \times 10^4$, and schedule $\tau$ at $40\%$ iterations for a period of $20\%$ iterations. Furthermore, to keep our sample weights consistent, we recover unscaled distance as $\sqrt{u(x)} / K$ before passing to $w_u$.

\begin{figure}[]
    \centering
    \includegraphics[width=\linewidth]{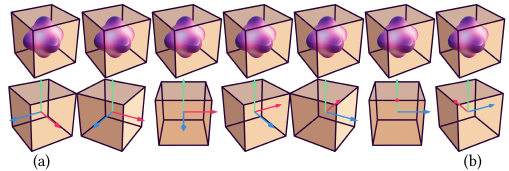}
    \caption{Consider 2 spatially adjacent points associated with the same octahedral frames under different rotation parameterization (a) and (b). Due to the continuity of MLP, frames queried in-between must be interpolated somehow (bottom row), yielding incorrect alignment of $\nabla u$. On the contrary, SH parameterization yields expected frames in between (top row). }
    \label{fig:sh-basis}
    \Description{Update ablation}
\end{figure}

\begin{figure}[]
    \centering
    \includegraphics[width=\linewidth]{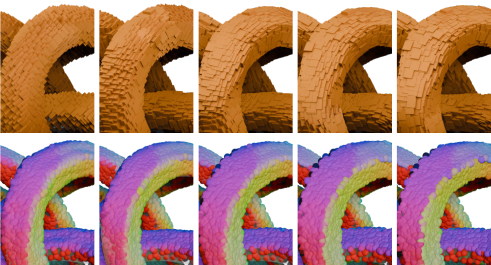}
    \caption{The update of octahedral frames at 5 iterations intervals (top row) and distance gradient at 300 iterations intervals (bottom row). The octahedral frames change from a nonsensical initial state to a reasonable alignment in merely 15 iterations.}
    \label{fig:update-ablation}
    \Description{Update ablation}
\end{figure}

\begin{figure*}[]
    \centering
    \includegraphics[width=\linewidth]{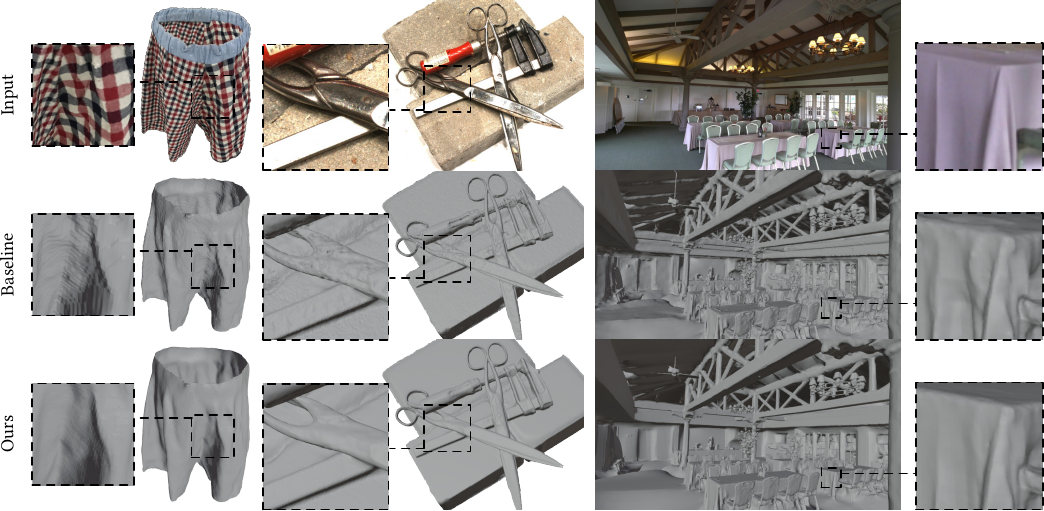}
    \caption{The first column shows results with NeuralUDF \cite{long2022neuraludflearningunsigneddistance} initialization, the second and third columns show results with Neuralangelo \cite{li2023neuralangelo} initialization. Although applicable for large-scale reconstructions, our current implementation lacks adaptive scaling that can smooth out details.}
    \label{fig:limitation2}
    \Description{neural rendering ablation}
\end{figure*}

\subsection{Result interpretation}
As shown in the full metrics (Tab. \ref{tab:full-metrics}), in the absence of input normals, our method achieves the best Chamfer distance at both noise levels on the ABC dataset, the best and second best Chamfer distance at noise level $0.002L$ and $0.01L$ on Thingi10k dataset, respectively. We also achieve the best F-score across two noise levels on both datasets. This clearly highlights our effectiveness in applying the octahedral field in reconstructing not only CAD-like, but also more general shapes from noisy unoriented point clouds.

As demonstrated in Fig. \ref{fig:limitation2}, our method is not tied to the medium of point cloud alone. The current limitation, the lack of adaptive scaling, is tied to our current formulation but not to the octahedral field itself. With proper scaling design and integration with the data prior, our method has the potential to be applied for scene-level surface reconstruction from raw images.

\section{Additional ablation studies}
\subsection{Alternative Octahedral Frame Representations}
The output of our octahedral MLP in $\mathbb{R}^9$ is not guaranteed to represent a valid octahedral frame. This raises two immediate questions: 1) Why not choose a parameterization that is guaranteed to output octahedral frames? and 2) How does it affect the update of $\nabla u$ using $\mathcal{L}_{\text{align}}$, if $q$ does not lie on the octahedral variety?

Theoretically, we could use the representation vector basis, or any form of rotation (e.g., rotvec, quaternion, rot6d \cite{Zhou_2019}, etc.) to parameterize the octahedral field, then evaluate equivalent smoothness in the SH parameterization space. However, since the same frame can be represented by multiple rotations, the smoothness in the octahedral space can lead to discontinuity in its parameterization space, which cannot be satisfied by the differentiability of MLP (Fig. \ref{fig:sh-basis}). Therefore, SH parameterization is irreplaceable.

For the second concern, we safeguard against it with two design choices--the normalization of octahedral MLP output and the linear scheduling of $\tau$. Upon normalization, the solution space of our octahedral MLP is very close to the octahedral variety, so $q$ converges rapidly (Fig. \ref{fig:update-ablation}). 

Furthermore, $\nabla u$ is partially governed by $\mathcal{L}_{\text{recon}}$, so in the early iterations of scheduling, $\mathcal{L}_{\text{align}}$ has a minimal effect on $u$. In contrast, $\mathcal{L}_{\text{align}}$ and $\mathcal{L}_{\text{smooth}}$ are the only two losses that act on $q$, and their relative weight is preserved in the scheduling. Given how fast $q$ converges (several iterations as validated in Fig. \ref{fig:update-ablation}), by the time $\tau$ rises to a non-negligible level, $q$ has already lies on the octahedral variety that gives a meaningful update to $\nabla u$. This further explains why we still need $\mathcal{L}_{\text{recon}}$ after the initialization. We further complement it with stability and convergence analysis of our two losses in the Sec. \ref{sec:stability}.

Note that the explanation above does not apply to the cases where the alignment is lifted by smoothness. Our rationale is that a lifted frame is in the proximity of aligned frames, which is unlikely to deviate much from the octahedral variety under explicit smoothness enforcement.

\begin{figure}[]
    \centering
    \includegraphics[width=\linewidth]{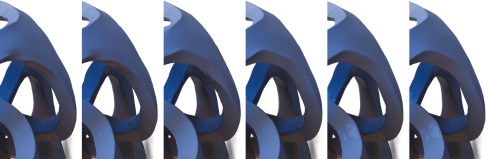}
    \caption{We set $\mathcal{L}_\text{align}=50$ then perturb  $\mathcal{L}_\text{lip}$ weight between $10^{-10}$ (left) to $1$ (right). Compared to $\mathcal{L}_\text{smooth}$, the alternative $\mathcal{L}_\text{lip}$ does not provide fine-grained control of reconstruction smoothness and sharpness.}
    \label{fig:lip-ablation}
    \Description{Lip ablation}
\end{figure}

\subsection{Alternative Smoothness Loss} \label{sec:lipmlp}
Given that the octahedral MLP output space is of Euclidean topology and is exactly where the smoothness is measured, an alternative and 
more efficient smoothness loss is the Lipschitz regularization. The LipMLP \cite{lipmlp} initializes the Lipschitz constant $c_i$ per layer to rescale the weight matrices
\begin{equation} \label{eq:lip-loss}
    \mathcal{L}_\text{lip} = \prod_{i=1}^l \text{softplus}(c_i).
\end{equation}
By shrinking all Lipschitz constants, the MLP's output variation with respect to input is minimized, hence is encouraged to be smooth. This loss is highly efficient because it only needs first-order derivatives, and all weight matrices can be updated in parallel.

In comparison to minimizing the gradient norm, however, we find that LipMLP lacks fine-grained control for our parameterization. An increase in Lipschitz regularization weight does not necessitate a smoother octahedral field (Fig. \ref{fig:lip-ablation}). Therefore, we choose the gradient norm as our smoothness loss to trade off interpretability over efficiency (Tab. \ref{tab:runtime}).

\begin{table}[]
    \caption{The runtime performance per scan. The LipMLP is more efficient, but at the cost of less control.}
    \label{tab:runtime}
    \begin{tabular}{ccc}
        \toprule
         \multicolumn{1}{c}{NSH}        & \multicolumn{1}{c}{Ours ($\mathcal{L}_{\text{smooth}}$) }             & \multicolumn{1}{c}{Ours ($\mathcal{L}_{\text{lip}}$) } \\ \hline
         7 min & 12 min & 10 min\\
        \bottomrule
    \end{tabular}
\end{table}

\section{Stability Analysis} \label{sec:stability}
We use the framework proposed in StEik \cite{yang2023steikstabilizingoptimizationneural}, which considers the gradient update of neural field as a geometric PDE, to analyze the stability of our proposed losses.

Specifically, for a neural field $u$ and a loss $L$
\begin{equation}
    L(u) = \int_\Omega f(u) dx,
\end{equation}
its evolution under gradient descent is governed by the negated functional derivative of $L$ over $u$, and can be expanded as
\begin{equation}
    \frac{\partial u}{\partial t} = - \frac{\delta L}{\delta u} = -\frac{\partial f}{\partial u} + \nabla \cdot \frac{\partial f}{\partial (\nabla u)},
\end{equation}
where $t$ is the continuum equivalent of the iteration index.

Therefore, its stability can be inferred from whether $\frac{\partial u}{\partial t}$ converges as $t \to \inf$.

\subsection{Smoothness} \label{sec:smooth-stable}
For smoothness loss, computing the functional derivative yields a forward diffusion process
\begin{equation}
    \frac{\partial q}{\partial t} = - \frac{\delta \mathcal{L}_{\text{smooth}}}{\delta q} = \nabla \cdot \frac{\partial}{\partial (\nabla q)}(\|\nabla q\|_F^2) = \nabla \cdot (2 \nabla q) = 2 \Delta q.
\end{equation}
Applying the Fourier Transform shows its solution converges as $t \to \inf$
\begin{equation}
    \frac{\partial \tilde{q}}{\partial t}(t, \omega) = -2\omega^2 \tilde{q}(t, \omega) \Rightarrow \tilde{q}(t, \omega) \propto e^{-2\omega^2 t}.
\end{equation}
Therefore, the gradient update of our smoothness loss is stable.

\subsection{Alignment}
For simplicity, we ignore normalization in the proof \cite{yang2023steikstabilizingoptimizationneural} by setting $r = \nabla u$. We then rewrite the gradient of the homogeneous polynomial as
\begin{equation}
    \nabla F_T(r, q) = c_0 c_1 J_0(r) + c_0 J_4(r)^T q \in \mathbb{R}^3,
\end{equation}
where $J_0(r) = \frac{d \hat{y}_0(r)}{d r} \in \mathbb{R}^3$, $J_4(r) = \frac{d \hat{y}_4(r)}{d r} \in \mathbb{R}^{9 \times 3}$.

\subsubsection{Alignment on $q$}
To analyze the stability of $\mathcal{L}_{\text{align}}$ on $q$, we treat $r$ as fixed vector locally and omit parentheses
\begin{equation}
    \mathcal{L}_{\text{align}}(q) = \int_{\partial \Omega} \|c_0 c_1 J_0 + c_0 J_4^T q - 4r\|_2^2 dx.
\end{equation}
Therefore
\begin{equation}
    \begin{aligned}
    \frac{\partial q}{\partial t} = -\frac{\delta \mathcal{L}_{\text{align}}}{\delta q} &= -\frac{\partial}{\partial q} \|c_0 c_1 J_0 + c_0 J_4^T q - 4r\|_2^2 \\
    &= -2 c_0 J_4(c_0 c_1 J_0 + c_0 J_4^T q - 4r) \\
    &= -2 c_0^2 J_4 J_4^T q + 2J_4(4c_0r - c_0^2 c_1J_0).
    \end{aligned}
\end{equation}
Since $J_4 J_4^T \in \mathbb{R}^{9 \times 9}$ is rank deficient, we can decompose $q$ as components on the column space and null space of $J_4 J_4^T$ respectively
\begin{equation}
    q = q_r + q_n, \ q_r \perp \text{null}(J_4 J_4^T), \ q_n \in \text{null}(J_4 J_4^T),
\end{equation}
so our analysis becomes
\begin{equation}
    \begin{aligned}
    \frac{\partial q}{\partial t} &= \frac{\partial q_r}{\partial t} + \frac{\partial q_n}{\partial t} \\
    \frac{\partial q_r}{\partial t} &= -2 c_0^2 J_4 J_4^T q_r + 2J_4(4c_0r - c_0^2 c_1J_0) \\
    \frac{\partial q_n}{\partial t} &= 0.
    \end{aligned}
\end{equation}
Note that $\frac{\partial q_r}{\partial t}$ is a first-order linear differential equation. Since $q_r^T (J_4 J_4^T) q_r > 0$, the restriction of the system matrix $-2 c_0^2 J_4 J_4^T$ on the column space of $J_4 J_4^T$ has all real and negative eigenvalues, yielding exponential stability. $\frac{\partial q_n}{\partial t}$ on the other hand, is marginally stable.

\begin{figure}[]
    \centering
    \includegraphics[width=\linewidth]{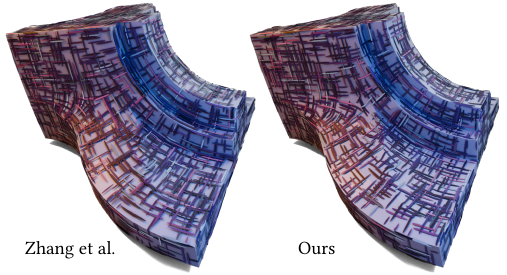}
    \caption{Our constraint produces an equivalent flowline as \citet{OctahedralAsCross} when applied for cross-field design.}
    \label{fig:flowline}
    \Description{NSH limitations.}
\end{figure}

The fixed points $q^*$ of $\frac{\partial q}{\partial t}$ satisfy the following normal equation
\begin{equation}
    J_4 J_4^T q^* = J_4(\frac{4}{c_0}r - c_1J_0) \\
    q^* = q_r^* + q_n, \ q_n \in \text{null}(J_4 J_4^T),
\end{equation}
with minimum norm solution $q_r^*$ also being the fixed point of $\frac{\partial q_r}{\partial t}$. Geometrically, $J_4^T q = \frac{4}{c_0}r - c_1J_0$ is equivalent normal alignment constraint in \cite{OctahedralAsCross}, but only 3 equations per frame as opposed to 7 (Fig. \ref{fig:flowline}). $q^*$ corresponds to the family of $r$-aligned frames with varying tangential twists, while $q_r^*$ represents $r$-aligned lobes of fixed norm $\sqrt{\frac{7}{12}}$ (see inset plots in Sec. 3.1 of \cite{OctahedralAsCross}).

Therefore, our loss pushes one direction of frame $q$ to align with $r$ at a linear rate of convergence, while allowing the other 2 to vary. Given that the $q$ is pre-normalized to a solution space very close to the true octahedral variety, $q$ should adapt very swiftly to the change of $r$, i.e., when $r$ shifts, $q$ quickly converges to the closest $r$-aligned octahedral frame.

The alignment loss itself does not constrain the tangential twists--the smoothness loss does.

\subsubsection{Alignment on $u$}
Similarly, we treat $q$ as locally fixed. Under the assumption that $q$ lies on the octahedral variety, we can use the original form of the homogeneous polynomial $F_T (r) = \sum_{i=1}^3 (v_i \cdot r)$, since its gradient is the same as that of $f_T$ under the parameterization of $q$
\begin{equation}
    \nabla F_T = \frac{\partial}{\partial r} c_0 (c_1 \hat{y}_0(r) + q^T \hat{y}_4(r)) = \frac{d}{d r} \sum_{i = 1}^3 (v_i \cdot r)^4 = 4 \sum_{i=1}^3 (v_i \cdot r)^3 v_i.
\end{equation}

For clarity, we start with notation $r = \nabla u$, so the alignment loss becomes
\begin{equation}
    \mathcal{L}_{\text{align}} (r) = \int_{\partial \Omega} \|4 \sum_{i=1}^3 (v_i \cdot r)^3 v_i - 4 r\|_2^2 dx,
\end{equation}
then
\begin{equation}
    \begin{aligned}
        \frac{\partial r}{\partial t} &= - \frac{\delta \mathcal{L}_{\text{align}}}{\delta r} \\
        &= - \frac{\partial}{\partial r} \|4 \sum_{i=1}^3 (v_i \cdot r)^3 v_i - 4 r\|_2^2 \\
&= -32 \cdot (3\sum_{i=1}^3 (v_i \cdot r)^5 v_i - 4 \sum_{i=1}^3(v_i \cdot r)^3 v_i + r).
    \end{aligned}
\end{equation}
Under $|r| = 1$, its fixed points are $r = \pm v_k$, $k \in \{1, 2, 3\}$.

Since $r = \nabla u$, with the product rule of divergence, we have
\begin{equation}
    \begin{aligned}
    \frac{\partial u}{\partial t} =& \nabla \cdot \frac{\partial}{\partial (\nabla u)} \|4 \sum_{i=1}^3 (v_i \cdot \nabla u)^3 v_i - 4 \nabla u\|_2^2 \\
    =& 32 \nabla \cdot (3\sum_{i=1}^3 (v_i \cdot \nabla u)^5 v_i - 4 \sum_{i=1}^3(v_i \cdot \nabla u)^3 v_i + \nabla u) \\
    =& 32 (3 \sum_{i=1}^3 (5(v_i \cdot \nabla u)^4 v_i^T H(u) v_i + (v_i \cdot \nabla u)^5(\nabla \cdot v_i)) \\
    &- 4 \sum_{i=1}^3 (3(v_i \cdot \nabla u)^2 v_i^T H(u) v_i + (v_i \cdot \nabla u)^3(\nabla \cdot v_i)) + \Delta u),
    \end{aligned} 
\end{equation}
where the Hessian $H(u)$ comes from the relation $\nabla(v_i \cdot \nabla u) = H(u) v_i$. With singular Hessian constraint $H(u) \nabla u = 0$ \cite{wang2023neuralsingularhessian}, it is trivial to verify that $\nabla u = \pm v_k$ remain the fixed points.

Denote fixed points $u_0$ such that $\nabla u_0 = \pm v_k$, we have
\begin{equation}
    v_i \cdot \nabla u_0 =
    \begin{cases}
    \pm 1 & i = k \\
    0 & i \neq k
    \end{cases}
\end{equation}
We add a small perturbation $\epsilon$ as
\begin{equation}
    u(t, x) = u_0(x) + \epsilon (t, x).
\end{equation}
From eikonal constraint $|\nabla u| = 1$ we derive
\begin{equation}
    |\pm v_k + \nabla \epsilon| = 1 \Rightarrow \nabla \epsilon \cdot v_k = 0 \Rightarrow H(\epsilon) v_k = \nabla (v_k \cdot \nabla \epsilon) = 0.
\end{equation}
Thus, under the assumption that $v_i$ is locally constant, we arrive at
\begin{equation}
    \frac{\partial u}{\partial t} = \frac{\partial \epsilon}{\partial t} = 32 \Delta \epsilon.
\end{equation}
Apply the same Fourier Transform as in Sec. \ref{sec:smooth-stable} shows the perturbation decays exponentially over time, suggesting our fixed points are stable.

\end{document}